\documentclass[final,5p,times,twocolumn]{elsarticle}
\usepackage{graphicx}
\usepackage{longtable}
\usepackage{multirow}
\usepackage{lineno,hyperref}
\usepackage{algpseudocode}
\usepackage[linesnumbered, ruled]{algorithm2e}
\usepackage{amsmath}
\usepackage{amssymb}
\usepackage[utf8]{inputenc}
\journal{X}
\usepackage{adjustbox}
\usepackage{float}
\usepackage{multirow}
\usepackage{makecell}
\usepackage{adjustbox}
\usepackage{multirow}
\usepackage{hyperref}

\hypersetup{
  colorlinks=true,
  linkcolor=blue,
  citecolor=blue,
  urlcolor=blue,
}

\begin{document}

\begin{frontmatter}

\title{ResCap-DBP: A Lightweight Residual-Capsule Network for Accurate DNA-Binding Protein Prediction Using Global ProteinBERT Embeddings}

\author[mymainaddress]{Samiul~Based~Shuvo\corref{cor1}}
\ead{sbshuvo@bme.buet.ac.bd}

\author[mymainaddress]{Tasnia~Binte~Mamun}
\ead{tasnia@bme.buet.ac.bd}

\author[mymainaddress1]{U~Rajendra~Acharya
}
\ead{Rajendra.Acharya@usq.edu.au}

\cortext[cor1]{Corresponding author}

\address[mymainaddress]{Department of Biomedical Engineering, Bangladesh University of Engineering and Technology (BUET), Dhaka-1205, Bangladesh}
\address[mymainaddress1]{School of Mathematics, Physics and Computing, University of Southern Queensland, Springfield, QLD 4300, Australia}

  \begin{abstract}
DNA-binding proteins (DBPs) are integral to gene regulation and cellular processes, making their accurate identification essential for understanding biological functions and disease mechanisms. Experimental methods for DBP identification are time-consuming and costly, driving the need for efficient computational prediction techniques. In this study, we propose a novel deep learning framework, ResCap-DBP, that combines a residual learning-based encoder with a one-dimensional Capsule Network (1D-CapsNet) to predict DBPs directly from raw protein sequences. Our architecture incorporates dilated convolutions within residual blocks to mitigate vanishing gradient issues and extract rich sequence features, while capsule layers with dynamic routing capture hierarchical and spatial relationships within the learned feature space. We conducted comprehensive ablation studies comparing global and local embeddings from ProteinBERT and conventional one-hot encoding. Results show that ProteinBERT embeddings substantially outperform other representations on large datasets. Although one-hot encoding showed marginal advantages on smaller datasets, such as PDB186, it struggled to scale effectively. Extensive evaluations on four pairs of publicly available benchmark datasets demonstrate that our model consistently outperforms current state-of-the-art methods. It achieved AUC scores of 98.0\% and 89.5\% on PDB14189 and PDB1075, respectively. On independent test sets PDB2272 and PDB186, the model attained top AUCs of 83.2\% and 83.3\%, while maintaining competitive performance on larger datasets such as PDB20000. Notably, the model maintains a well-balanced sensitivity and specificity across datasets. These results demonstrate the efficacy and generalizability of integrating global protein representations with advanced deep learning architectures for reliable and scalable DBP prediction in diverse genomic contexts.

    \end{abstract}
    
    
    \end{frontmatter}
    
    \section{Introduction}

DNA-binding proteins (DBPs) exhibit a unique affinity for DNA molecules, playing crucial roles in essential cellular processes such as gene regulation, replication, repair, and transcriptional control \cite{ahmed2018integrated,sandman1998diversity,latchman1997transcription}. Approximately 6- 7\% of eukaryotic genomes and 2-5\% of prokaryotic genomes encode DBPs. The precise identification of DBPs is crucial in biological research and drug development, particularly due to their complex interactions with DNA and their involvement in regulatory pathways associated with various diseases \cite{chowdhury2017idnaprot}. Traditionally, DBP identification has heavily depended on experimental methods such as nuclear magnetic resonance (NMR) spectroscopy \cite{khalil2023nuclear},and chromatin immunoprecipitation (ChIP) assays and  X-ray crystallography \cite{srivastava2018role}. While these experimental techniques offer high accuracy, they are typically labor-intensive, time-consuming, costly, and impractical for high-throughput analyses.

To overcome these limitations, computational methodologies that leverage machine learning and deep learning have become increasingly prevalent and effective alternatives. These computational techniques can rapidly predict the status of DBP directly from protein sequences or structural data, thereby significantly enhancing both efficiency and scalability. Broadly, computational approaches fall into two main categories: sequence-based  and structure-based methods. Structure-based methods, such as DBD-Hunter \cite{gao2008dbd} and REGAd3 \cite{iqbal2015improved},  utilize detailed three-dimensional protein structural information to predict DBPs. Although these methods often yield accurate predictions, their effectiveness is frequently constrained by the limited availability of resolved protein structures. In contrast, sequence-based methods, including DNAbinder \cite{raghava2007identification}, SVM-PSSM \cite{ho2007design}, DNA-BIND-PROT \cite{ozbek2010dnabindprot}, DR-bind \cite{chen2012dr_bind}, and iDNAPro-PseAAC \cite{liu2015dna}, primarily depend on sequence-derived features. 

These methods, although often more practical due to broader data availability, typically involve intricate feature engineering and manual feature selection, which increases computational complexity and limits their applicability to large-scale datasets. Recent advances in deep learning have significantly improved the predictive capabilities for DBP identification by automating feature extraction processes and capturing complex patterns in data. Qu et al. \cite{qu2017prediction} effectively combined Long Short-Term Memory (LSTM) networks and one-dimensional convolutional neural networks (CNNs) to predict DBPs directly from raw amino acid sequences, thereby eliminating the need for manual feature selection. Hassanzadeh et al. \cite{hassanzadeh2016deeperbind}  introduced an integrated architecture comprising LSTM and two-dimensional CNN layers, which achieves enhanced prediction accuracy through the combined learning of sequential and spatial features. Similarly, Barukab et al. \cite{barukab2022dbp} proposed DBP-CNN, which leverages evolutionary protein properties to enhance classification performance. Mursalim et al.\cite{mursalim2023bicaps} developed BiCaps-DBP, integrating bidirectional LSTM (Bi-LSTM) and Capsule Networks (CapsNet), addressing previous limitations in predictive specificity. Moreover, Ahmed et al.\cite{ahmed2024stackdpp}  critically evaluated widely used benchmark datasets, such as PDB1075 and PDB186, revealing overlaps and sequence similarity biases that compromised generalization claims. To address this, they proposed new datasets UNIPROT1424 and UNIPROT356 with rigorous filtering criteria and lower inter-dataset homology. Their model, StackDPP, employed recursive feature elimination and stacking ensemble learning to achieve high performance. However, it depended on hundreds of complex features and a multistage pipeline, restricting its use in end-to-end learning scenarios and large-scale applications. Zeng et al. \cite{zeng2024lbi} proposed LBi-DBP, a lightweight BiLSTM model with attention mechanisms to offer interpretable predictions. Although efficient, it still inherits the limitations of LSTMs in modeling long-range dependencies and scalability to ultra-long sequences. In parallel, Zhang et al.\cite{zhang2024drbppred}  introduced DRBPPred-GAT, a residue-level predictor that utilizes graph attention networks and stacked autoencoders for feature enhancement. Although it achieved state-of-the-art performance, its heavy reliance on preprocessing, structural data, and multiple feature embeddings compromises simplicity and end-to-end applicability. Overall, while these studies represent significant progress, major limitations persist: overfitting due to training-test overlap, excessive computational complexity, manual feature engineering, interpretability trade-offs, and impracticality for real-time deployment remain unresolved challenges. These challenges highlight the need for models that strike a balance between simplicity, generalizability, interpretability, and computational efficiency. Traditional CNNs often struggle to capture nuanced correlations between features, while LSTMs face issues such as vanishing gradients and high computational costs. To overcome these limitations, Oord et al. \cite{oord2016wavenet} incorporated residual learning modules into WaveNet architectures, significantly improving model convergence and reducing training time. Additionally, Capsule Networks \cite{hinton2018matrix} introduced dynamic routing algorithms to capture hierarchical feature relationships more effectively, thereby enhancing robustness and predictive performance.

\begin{figure*}[!htbp]
        \centering
        \includegraphics[width=\linewidth]{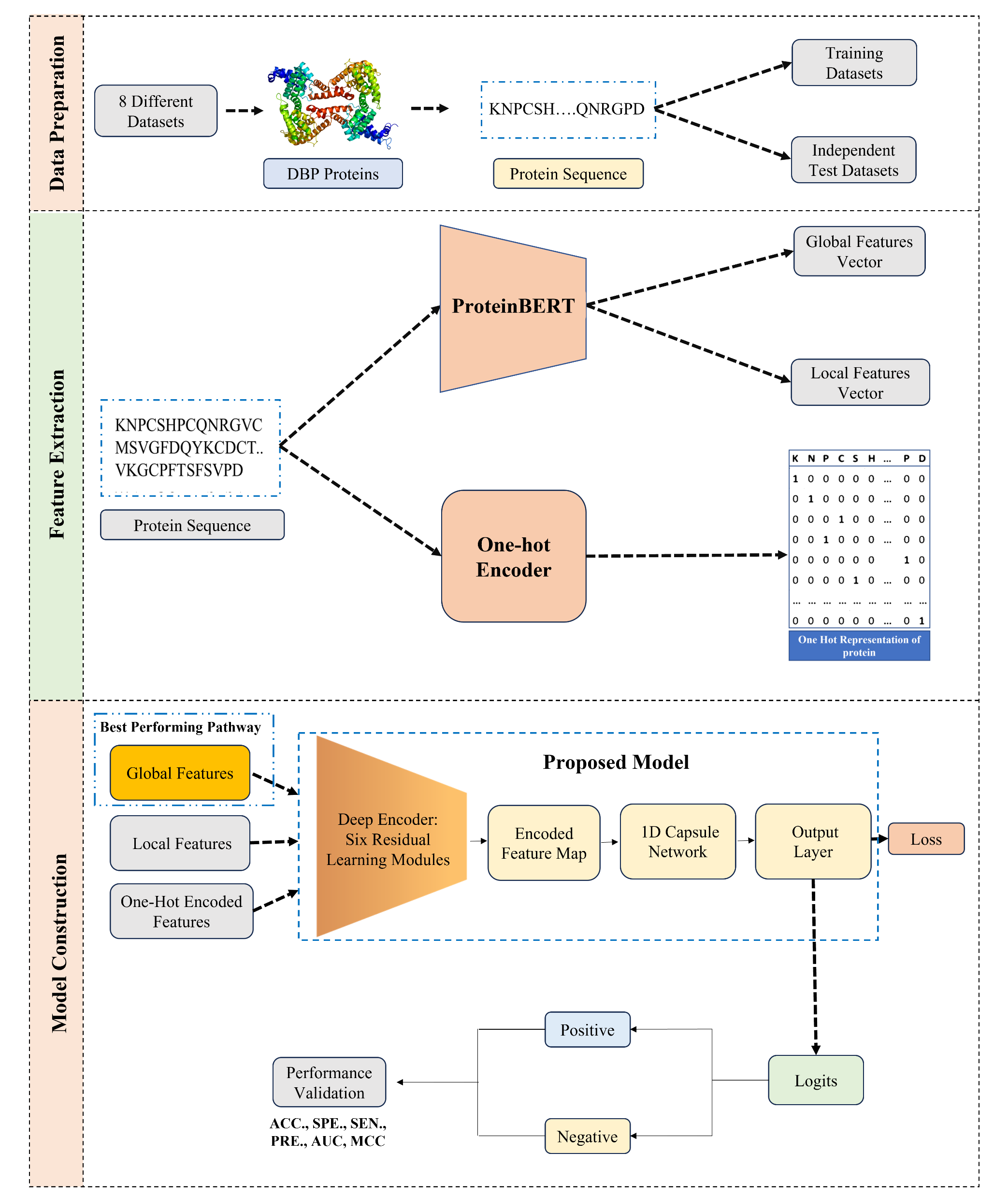}
        \caption{Graphical abstract of the proposed DNA-Binding Protein (DBP) prediction framework. The Pipeline involves three main stages: (1) Data Preparation: Protein sequences from eight benchmark datasets are divided into Training and Independent test sets; (2) Feature Extraction: Protein Sequences are encoded using ProteinBert for Global And Local contextual features, and One-hot encoding for Residue-level representation; (3) Model Construction: The Proposed deep learning model integrates Global Features into a Residual learning-based encoder followed by a 1D Capsule network to produce final predictions. Model performance is evaluated using standard classification metrics including Accuracy (ACC.), Sensitivity (SEN.), Specificity (SPE.), Precision (PRE.), AUC, And MCC.}
        \label{GA}
\end{figure*}

Building upon these advancements, we propose a novel deep learning architecture, ResCap-DBP, that synergistically combines residual learning and Capsule blocks to model complex spatial and contextual dependencies in biological sequence data. As illustrated in the graphical abstract (Figure~\ref{GA}), the architecture is designed to be both lightweight, enabling efficient training and inference without compromising accuracy.

    The contributions of our work are summarized as follows:
    \begin{enumerate}
        \item We introduced a novel deep-learning architecture(ResCap-DBP) designed to predict DBPs in real time. Our proposed model has demonstrated superior performance compared to existing state-of-the-art (SOTA) architectures.

        \item We conducted a comprehensive feature selection analysis by comparing the performance of one-hot encoded features with the global and local features extracted by ProteinBERT \cite{brandes2022proteinbert}. Our findings revealed that global features of ProteinBERT achieved the highest performance-to-complexity ratio.
        \item We conducted extensive benchmarking using four distinct pairs of data to evaluate the effectiveness of our deep learning architecture rigorously. Each set comprises an independent training and test set, allowing for a comprehensive and robust assessment of our model’s capabilities. 
        \item We optimized our proposed network for low processing power and minimal inference time, making it inherently lightweight and efficient. Its ability to deliver high throughput with reduced computational overhead enhances its practicality for real-world applications, especially in resource-constrained environments.

    \end{enumerate}
In Section \ref{two}, we present a detailed overview of the datasets used and the methodological framework. Section \ref{three} describes the proposed network architecture and its main components. Section \ref{four} outlines the experimental setup and presents quantitative results comparison, followed by a discussion of their significance. In Section \ref{five}, we explore the study’s limitations and suggest directions for future research. The paper ends with Section \ref{six}, which recaps the key contributions

    \section{Datasets}\label{two}
\begingroup
\setlength{\tabcolsep}{2pt} 
\renewcommand{\arraystretch}{1.35} 
\begin{table}[t]
\centering
\caption{Summary of datasets used in this study}
\label{tab:dataset_overview}
\scriptsize 
\begin{tabular}{p{1.1cm}|c|c|c|c|c}
\hline\hline
\multirow{2}{*}{References} & \multirow{2}{*}{Datasets} & \multicolumn{2}{c|}{Training Set} & \multicolumn{2}{c}{Test Set} \\
\cline{3-6}
& & DBPs & Non-DBPs & DBPs & Non-DBPs \\
\hline\hline
\cite{barukab2022dbp, du2019msdbp, zou2013improved, ma2016dnabp} & PDB14189 / PDB2272 & 7129 & 7060 & 1153 & 1119 \\
\hline
\cite{raghava2007identification, lin2011idna, kumar2009dna, szilagyi2006efficient} & PDB1075 / PDB186 & 525 & 550 & 93 & 93 \\
\hline
\cite{hu2019improved} & PDB67151 / PDB20000 & 17,151 & 50,000 & 10,000 & 10,000 \\
\hline
\cite{ahmed2022sl} & UNIPROT1424 / UNIPROT356 & 712 & 712 & 178 & 178 \\
\hline\hline
\end{tabular}
\end{table}
\endgroup
    \begin{figure*}[!htbp]
        \centering
        \includegraphics[width=\linewidth]{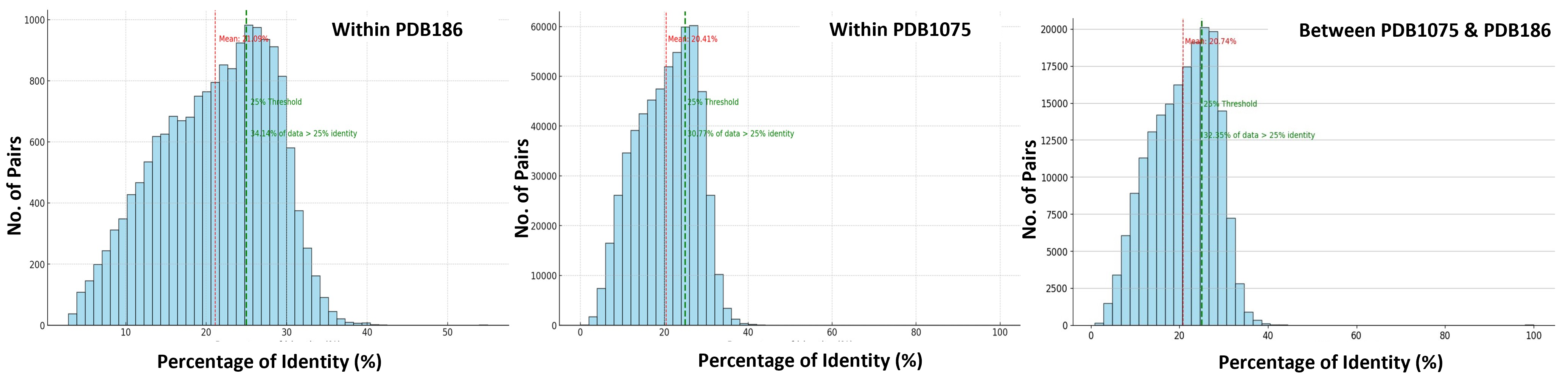}
        \caption{Pairwise percentages of identity calculated using the Needleman-Wunsch algorithm:(a) within PDB186 ,(b)within PDB1075 and (c) between PDB1075 and PDB186.}
        \label{fig:data_reliability}
    \end{figure*}

The present study evaluates model generalization and robustness using four benchmark training–test dataset pairs derived from eight publicly available sources (Table~\ref{tab:dataset_overview}). The first pair, PDB14189/PDB2272, comprises 7,129 DNA‐binding proteins (DBPs) and 7,060 non‐DBPs for training, with 1,153 DBPs and 1,119 non‐DBPs reserved for independent testing. The second pair, PDB1075/PDB186, includes 525 DBPs and 550 non‐DBPs in the training set, with 93 DBPs and 93 non‐DBPs for testing. Although our experimental evaluations were conducted across all datasets, we restricted detailed reliability analysis, specifically, pairwise sequence similarity assessment, to PDB1075 and PDB186 due to the substantial computational cost of large-scale sequence alignment. Using the Needleman–Wunsch global alignment algorithm with the BLOSUM62 scoring matrix, we observed an average pairwise sequence identity of 20.41\% in PDB1075 and 21.09\% in PDB186, with approximately 30–34\% of pairs exceeding the 25\% identity threshold. Additionally, 32.35\% of inter-dataset sequence pairs between PDB1075 and PDB186 surpassed this threshold, and 42 identical sequences were found in both training and test sets. These findings, illustrated in Figure~\ref{fig:data_reliability}, raise concerns about potential redundancy and the risk of performance overestimation in these datasets. To address these quality concerns and mitigate potential data leakage, we also evaluated our models using UNIPROT1424/UNIPROT356, a dataset pair introduced by Ahmed et al. \cite{ahmed2022sl}, which explicitly removes overlapping and redundant sequences between training and test sets. It consists of 712 DBPs and 712 non-DBPs in the training split and 178 of each class for testing, ensuring non-redundant and more realistic generalisation assessment. The third benchmark, PDB67151/PDB20000, investigates performance under class imbalance, using 17,151 DBPs and 50,000 non‐DBPs for training, with a balanced test set of 10,000 examples per class. All datasets were curated following established protocols. For instance, PDB14189 sequences were filtered to lengths between 50 and 6,000 residues and reduced to less than 25\% pairwise similarity \cite{ma2016dnabp}. Overall, this diverse selection of balanced, imbalanced, and quality-controlled datasets supports a rigorous and comprehensive evaluation of our model’s performance.

    \section{Proposed Network Architecture}\label{three}
    \subsection{Network fundamentals}
    \begin{figure*}[!h]
        \centering    \includegraphics[width=1\linewidth]{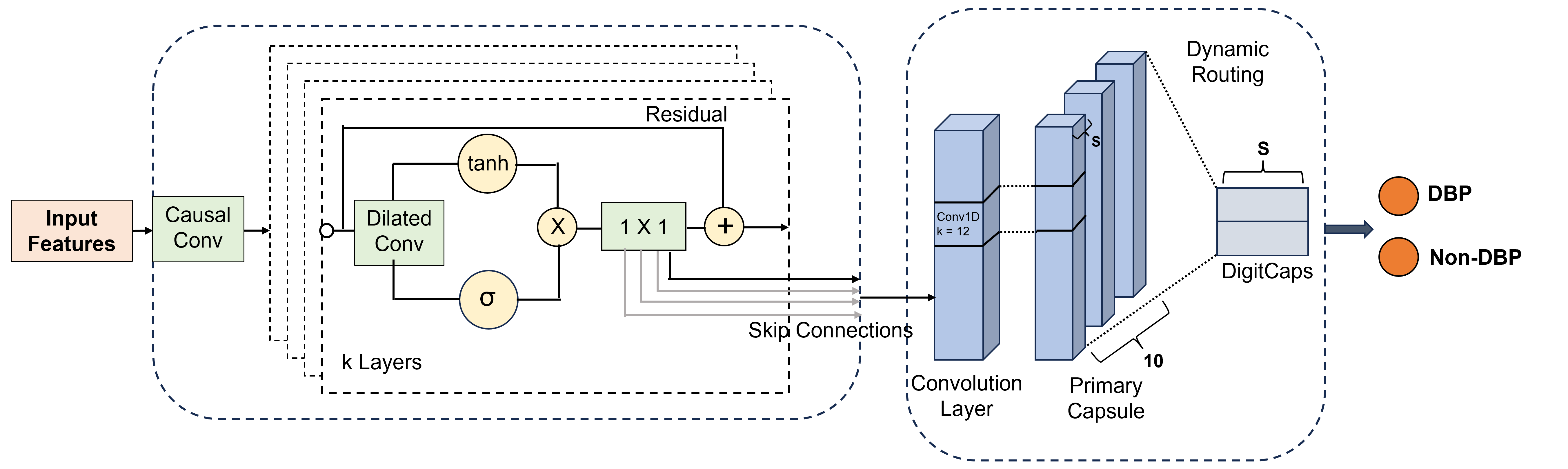}
        \caption{Overview of the proposed ResCap-DBP architecture. It combines Causal and Dilated Convolutions with Residual and Skip connections to capture Long-range dependencies, followed by a Capsule network for Dynamic feature routing and final prediction of DNA-Binding Proteins. }
        \label{fig:model}
    \end{figure*}
\subsubsection{Residual Learning Module:}  
Deep neural networks often face difficulties in training as the number of layers increases, primarily due to the vanishing gradient problem. To alleviate this, WaveNet \cite{oord2016wavenet} integrates a residual learning framework, which facilitates more effective gradient propagation and accelerates convergence. A central component of this architecture is the use of \textit{dilated convolutional layers}, which introduce gaps (dilations) between filter elements, enabling the network to exponentially expand its receptive field without increasing the number of parameters. This mechanism allows the model to capture long-range dependencies in the data while maintaining computational efficiency. Within each residual block, 1$\times$1 convolutions are employed to transform feature representations while preserving the input's spatial dimensions. Additionally, \textit{shortcut connections} are used to perform element-wise summation between the input and the output of the residual layers. These connections not only support gradient flow across deeper layers but also allow the network to learn identity mappings, which are critical for stabilizing and deepening the model architecture. Collectively, the residual module enhances both training stability and representational power.

\subsubsection{1D-CapsNet:}  
Capsule Networks (CapsNets), introduced by Hinton et al. \cite{hinton2018matrix}, were designed to overcome fundamental limitations of Convolutional Neural Networks (CNNs), particularly their inability to adequately model spatial hierarchies and part-whole relationships. CNNs, while effective at feature extraction, rely heavily on pooling operations, which can lead to the loss of important spatial information and hinder generalization to variations in input pose or orientation. In contrast, CapsNets represent features as \textit{vectors} (capsules) rather than scalar activations, capturing both the presence and properties (e.g., orientation, scale) of detected patterns. A key innovation in CapsNet is the \textit{dynamic routing algorithm}, which replaces traditional pooling by allowing capsules to selectively route their outputs to higher-level capsules based on agreement—thereby learning the part-to-whole relationships more effectively. When adapted to one-dimensional data (1D-CapsNet), such as biological sequences or temporal signals, this architecture demonstrates superior capability in preserving feature structure and enhancing interpretability. By modeling inter-feature dependencies and spatial configurations more faithfully, 1D-CapsNet holds significant promise for improving accuracy in sequence-based tasks such as protein function annotation or genomic classification.

\subsection{Proposed architecture}  
We proposed ResCap-DBP, a deep learning architecture that integrates Residual Learning with a one-dimensional Capsule Network (1D-CapsNet) to enable robust and accurate identification of DNA-binding proteins (see Figure~\ref{fig:model}. To extract both local and global contextual features, the input matrix \(\mathbf{X}\) is processed through a deep encoder comprising six Residual Learning Modules. Each module includes dilated convolutional layers, which allow the network to expand its receptive field without increasing the number of parameters or computational complexity. The operation of each module is expressed as:
\[
\mathbf{R}^{(l)} = \mathbf{X}^{(l-1)} + \mathcal{F}(\mathbf{X}^{(l-1)}, \mathbf{W}^{(l)}),
\]
where \(\mathbf{X}^{(l-1)}\) is the input to the \(l\)-th residual block, \(\mathbf{R}^{(l)}\) is the output, and \(\mathcal{F}(\cdot)\) is the residual function comprising dilated convolutions with dilation rate \(r_l\), pointwise \(1 \times 1\) convolutions, batch normalization, and nonlinear activations. The identity shortcut connection \(\mathbf{X}^{(l-1)}\) promotes stable gradient flow, enabling deeper network training without degradation.  The dilation rates in successive layers are typically set to increase exponentially (e.g., \(r_l = 2^{l-1}\)) to capture multiscale sequence dependencies. The final output of the encoder, denoted \(\mathbf{R}^{(6)}\), is a rich, hierarchical feature representation of the protein sequence. This encoded feature map is then passed into a 1D-CapsNet layer. The capsule network employs vectorized capsules and dynamic routing to model part-whole relationships between features. Unlike traditional convolutional layers, capsule networks preserve spatial hierarchies and can capture intricate patterns essential for robust classification. By combining the deep, context-aware feature extraction of residual modules with the structural modeling capacity of capsule networks, the proposed architecture effectively identifies DNA-binding proteins. This synergy allows the model to learn both fine-grained local features and higher-order global interactions that are critical for accurate classification. The architecture is designed to leverage the strengths of residual learning and capsule networks for robust, hierarchical, and scalable modeling of biological sequences.

To further validate our architectural choices, we tested multiple simplified variants of ResCap-DBP in Section~\ref{abl}. 

\section{ Experimental Setups }\label{four}

This study employed a rigorous evaluation protocol by maintaining a dedicated independent test set, which remained entirely unseen during both training and hyperparameter tuning. Model selection and validation were performed using five-fold cross-validation exclusively on the training data, ensuring robust assessment while preserving the integrity of the test set. This strategy provides an unbiased and accurate estimate of the model’s generalization performance. 
Given the imbalanced nature of DNA-binding protein datasets, the performance evaluation of the method was conducted using several standard metrics: accuracy (ACC), precision (PRE), sensitivity (SEN), specificity (SPE), Matthews correlation coefficient (MCC), and the area under the ROC curve (AUC). Definitions of these metrics are provided below:

The overall classification accuracy (Acc.) is calculated as:
\begin{equation}
    \text{ACC.} = \frac{TP + TN}{TP + TN + FP + FN}
\end{equation}

Precision (PRE.), also known as the positive predictive value, is given by:
\begin{equation}
    \text{PRE.} = \frac{TP}{TP + FP}
\end{equation}

 Sensitivity (SEN.), measures the proportion of actual positives correctly identified:
\begin{equation}
    \text{SEN.} = \frac{TP}{TP + FN}
\end{equation}

Specificity (SPE.) evaluates the proportion of actual negatives correctly identified:
\begin{equation}
    \text{SPE.} = \frac{TN}{TN + FP}
\end{equation}

The Matthews correlation coefficient (MCC), a robust metric particularly suitable for imbalanced datasets, is given by:
\begin{equation}
    \text{MCC} = \frac{TP \cdot TN - FP \cdot FN}{\sqrt{(TP + FP)(TP + FN)(TN + FP)(TN + FN)}}
\end{equation}

Finally, the area under the receiver operating characteristic (ROC) curve (AUC) evaluates the model's capability to distinguish between positive and negative classes over a range of decision thresholds.

Here, \( TP \) (true positives) refers to DNA-binding proteins correctly predicted as binding; \( TN \) (true negatives) denotes non-binding proteins correctly identified as non-binding. Conversely, \( FP \) (false positives) are non-binding proteins incorrectly predicted as binding, while \( FN \) (false negatives) are DNA-binding proteins incorrectly classified as non-binding.

All experiments were carried out using the Keras framework and executed on a computational server with NVIDIA A100 GPUs (80 GB VRAM). Hyperparameter tuning was performed empirically. A batch size of 128 was chosen to ensure a balance between convergence and efficiency. The number of routing iterations in the Capsule Network was fixed at 2 to promote consistency in capsule agreement. Binary cross-entropy was selected as the loss function, as it is well-suited for binary classification tasks.

    \section{Results and Discussion}\label{five}
\subsection{Ablation study on architectural components of ResCap-DBP}\label{abl}

To examine the individual contributions of ResCap-DBP’s core components, we conducted an ablation study using the UNIPROT1424 dataset. We designed five baseline models by systematically altering one major architectural element at a time and compared their performance with that of the full proposed model. Baseline 1, a stripped-down model that removes both the capsule layer and the residual encoder, replacing them with a vanilla CNN comprised of a few stacked 1D convolutional layers, and takes the global ProteinBERT embeddings. Baseline 2 removes the capsule network while retaining the global ProteinBERT embeddings and residual dilated convolutional encoder, using a simple fully connected layer for classification. Baseline 3 excludes the residual learning modules and passes the ProteinBERT global embeddings directly into a 1D capsule network. Baseline 4 retains the full ResCap-DBP architecture but replaces the ProteinBERT embeddings with traditional one-hot encoding. Baseline 5 also retains the full architecture but uses local residue-level embeddings from ProteinBERT instead of global contextual ones. The complete ResCap-DBP model incorporates global ProteinBERT embeddings, residual blocks, and 1D capsule layers for learning hierarchical representations.

The comparative results in Table~\ref{tab:ablation_uniprot} demonstrate the distinct impact of each architectural component on the overall performance of the ResCap-DBP model. The removal of both the residual encoder and capsule network in Baseline 1 yields the most basic configuration, resulting in relatively lower accuracy (84.7\%) and MCC (67.9\%). When only the capsule network is excluded (Baseline 2), performance improves notably (87.6\% accuracy, 75.3\% MCC), indicating that the residual encoder contributes more significantly to model effectiveness than capsule routing. Replacing the residual encoder with capsule layers (Baseline 3) results in slightly reduced performance (86.2\% accuracy, 72.5\% MCC) compared to Baseline 2. This highlights that while capsule networks are capable of modeling spatial hierarchies, they are less effective than residual pathways in capturing relevant sequence patterns on their own. The strong performance of Baseline 2 relative to Baseline 3 underscores the importance of residual learning in enabling deeper and more stable feature extraction. Baseline 4 exhibits the most severe performance degradation (79.0\% accuracy, 58.4\% MCC), despite retaining the full model architecture, due to the substitution of contextual ProteinBERT embeddings with one-hot encodings. This result strongly affirms the critical role of pretrained, global protein sequence representations in learning discriminative features. Baseline 5, utilizing local residue-level embeddings from ProteinBERT, achieves high sensitivity (96.1\%) but exhibits considerably lower specificity (69.1\%) and MCC (63.8\%), suggesting overfitting towards the positive class and a limited ability to generalize across negative samples. This further emphasizes that global embeddings, capturing long-range dependencies, are better suited for this classification task.

In contrast, the complete ResCap-DBP model outperforms all baselines across every metric, achieving 91.1\% accuracy, 94.1\% sensitivity, 88.1\% specificity, and an MCC of 82.3\%. These improvements validate the synergistic integration of global contextual embeddings, residual convolutional encoders, and capsule-based feature modeling in achieving a well-balanced and highly generalizable DBP prediction framework.

\begingroup
\setlength{\tabcolsep}{4pt}
\renewcommand{\arraystretch}{1.4}
\begin{table}[ht]
\centering
\caption{Ablation Study on the Proposed ResCap-DBP Architecture.}
\resizebox{\columnwidth}{!}{
\begin{tabular}{lcccccc}
\hline \hline
\textbf{Model Variant} & \textbf{ACC. (\%)} & \textbf{SEN. (\%)} & \textbf{SPE. (\%)} & \textbf{PRE. (\%)} & \textbf{AUC (\%)} & \textbf{MCC (\%)} \\
\hline \hline
Baseline 1& 84.7 & 93.4 & 74.5 & 78.6 & 87.9 & 67.9 \\
Baseline 2& 87.6 & 91.0 & 84.2 & 84.7 & 90.1 & 75.3 \\
Baseline 3& 86.2 & 89.9 & 82.3 & 83.1 & 89.2 & 72.5 \\
Baseline 4& 79.0 & 91.6 & 66.3 & 71.3 & 85.2 & 58.4 \\
Baseline 5& 82.3 & 96.1 & 69.1 & 75.4 & 86.7 & 63.8 \\

\textbf{ResCap-DBP} & \textbf{91.1} & \textbf{94.1} & \textbf{88.1} & \textbf{88.7} & \textbf{91.1} & \textbf{82.31} \\
\hline \hline
\end{tabular}
}
\label{tab:ablation_uniprot}
\end{table}
\endgroup

    \subsection{Evaluation of the optimal feature extraction techniques}

To investigate the impact of feature representation on DBP classification
performance, we evaluated three distinct approaches: the traditional one-hot
encoding, local contextual embeddings from ProteinBERT, and global contextual embeddings from ProteinBERT. A comparative analysis was conducted across four widely used benchmark datasets: PDB14189, PDB2272, PDB1075, and PDB186, which represent diverse data scales and sequence characteristics (see Table~\ref{tab:comparison_feature_extraction}).

    \subsubsection{One-Hot encoding}
    In this study, protein sequences are converted into a numerical format. Each protein sequence of length \(L\) is mapped to a fixed‐size binary indicator matrix \(X \in \{0,1\}^{L_{\max}\times 20}\). We enumerate the 20 canonical amino acids in a fixed order (e.g., A, C, D, …, Y). For residue \(i\), the one‐hot vector
\[
X_{i,\,:} = [0,\dots,0,1,0,\dots,0]
\]
has a ‘1’ in the column corresponding to the amino acid at position \(i\) and zeros elsewhere. Sequences shorter than \(L_{\max}\) are right‐padded with a special “PAD” token (all‐zero vector), and sequences longer than \(L_{\max}\) are truncated. This encoding preserves exact residue identity without introducing trainable parameters. In our case, \(L_{\max}\) is equal to 1000.

\begingroup
\setlength{\tabcolsep}{4pt} 
\renewcommand{\arraystretch}{1.35} 

\subsubsection{ProteinBERT Tokenization and Embeddings}
We employed ProteinBERT~\cite{brandes2022proteinbert}, a transformer-based language model pre-trained on millions of protein sequences. Input sequences are first tokenized into \(L\) tokens (including special [CLS], [SEP], and [PAD] tokens). Each token is embedded into \(\mathbb{R}^{d_{\mathrm{emb}}}\) via a lookup table (\(d_{\mathrm{emb}}=768\)), and learnable positional encodings are added to capture order information.
\par ProteinBERT’s dual‐pathway architecture produces both residue‐level and sequence‐level features:
\begin{itemize}
  \item \textbf{Local Representations.} A tensor
  \[
    \mathbf{L} \in \mathbb{R}^{B\times L\times d_{\mathrm{local}}},
    \quad d_{\mathrm{local}} = 512,
  \]
  where \(B\) is the batch size and \(L\) the padded sequence length. Each slice \(\mathbf{L}_{b,i,:}\) captures contextualized information for residue \(i\).

  \item \textbf{Global Representations.} A matrix
  \[
    \mathbf{G} \in \mathbb{R}^{B\times d_{\mathrm{global}}},
    \quad d_{\mathrm{global}} = 512,
  \]
  obtained via attention‐pooling (e.g., the [CLS] token), summarizing the entire sequence.
\end{itemize}

\par In the large-scale data sets PDB14189 and PDB2272, the global ProteinBERT
embeddings achieved the most robust and balanced performance in all evaluation metrics. Specifically, on PDB14189, global embeddings achieved an accuracy of 95.01\%, a specificity of 90.42\%, a precision of 91.11\%, an AUC of 98.00\%, and an MCC of 88.18\%, outperforming both one-hot and local representations by significant margins. A similar trend was observed on PDB2272, where global embeddings yielded the highest accuracy (79.29\%), specificity (65.14\%), precision (70.76\%), AUC (83.20\%), and MCC (50.21\%). On smaller datasets such as PDB1075 and PDB186, the performance advantage of global embeddings was less pronounced, though still evident in several metrics. For instance, on PDB1075, ProteinBERT (Global) achieved the highest specificity (65.14\%), precision (70.76\%), and AUC (89.50\%). Likewise, on PDB186, it secured the best specificity (46.41\%), AUC (83.30\%), and MCC (34.40\%). However, one-hot encoding remained competitive on these datasets, particularly in terms of MCC, achieving 46.58\% on PDB1075 and 32.39\% on PDB186.

\begingroup
\setlength{\tabcolsep}{4pt} 
\renewcommand{\arraystretch}{1.35} 
\begin{table}[t]
\centering
\caption{Comparison of different feature extraction techniques}
\label{tab:comparison_feature_extraction}

\resizebox{\columnwidth}{!}{%
\begin{tabular}{c|c|c|c|c|c|c}
\hline \hline
\multirow{2}{*}{\textbf{Feature Representation}} & \textbf{ACC. (\%)} & \textbf{SEN. (\%)} & \textbf{SPE. (\%)} & \textbf{PRE. (\%)} & \textbf{AUC. (\%)} & \textbf{MCC. (\%)} \\
\cline{2-7}
 & \multicolumn{6}{c}{\textbf{PDB14189}} \\
\hline \hline
One-Hot Encoding & 76.44 & 98.34 & 52.82 & 73.43 & 78.29 & 51.44 \\
ProteinBert (Local) & 79.01 & \textbf{98.62} & 22.54 & 56.37 & 74.24 & 20.42 \\

ProteinBert (Global) & \textbf{95.01} & 97.43 & \textbf{90.42} & \textbf{91.11} & \textbf{98.00} & \textbf{88.18} \\
\hline \hline
 & \multicolumn{6}{c}{\textbf{PDB2272}} \\
\hline \hline
One-Hot Encoding & 66.13 & 90.86 & 39.03 & 62.49 & 69.76 & 30.57 \\
ProteinBert (Local) & 70.79 & \textbf{97.45} & 16.12 & 54.10 & 72.78 & 12.49 \\

ProteinBert (Global)  & \textbf{79.29} & 83.99 & \textbf{65.14} & \textbf{70.76} & \textbf{83.20} & \textbf{50.21} \\
    \hline
\hline
    
 & \multicolumn{6}{c}{\textbf{PDB1075}} \\
\hline \hline
One-Hot Encoding & 78.88 & 97.30 & 49.79 & 68.88 & 84.01 & \textbf{50.38} \\
ProteinBert (Local) & 76.14 & \textbf{99.77} & 21.91 & 56.32 & 76.26 & 14.45 \\
    ProteinBert (Global) & \textbf{79.29} & 83.99 & \textbf{65.14} & \textbf{70.76} & \textbf{89.5} &  50.21 \\
\hline \hline
 & \multicolumn{6}{c}{\textbf{PDB186}} \\
\hline \hline
One-Hot Encoding & \textbf{68.02} & 89.45 & 43.02 & \textbf{64.77} & 71.56 & 32.39 \\
ProteinBERT (Local) & 66.67 & \textbf{99.63} & 15.18 & 53.80 & 68.04 & 17.53 \\
ProteinBERT (Global) & 67.71 & 84.94 & \textbf{46.41} & 61.66 & \textbf{83.30} & \textbf{34.40} \\
\hline \hline
\end{tabular}
}

\end{table}
\endgroup
On the other hand, ProteinBERT’s local embeddings demonstrated consistently high sensitivity across all datasets, often exceeding 97\%. While this indicates that the local context encoder is highly effective at detecting positive class instances, it comes at the cost of low specificity, precision, and MCC. For example, on PDB2272, the local embedding achieved a sensitivity of 97.45\% but a specificity of only 16.12\% and an MCC of 12.49\%. This imbalance indicates a strong positive-class bias, suggesting that local embeddings lack the holistic sequence information necessary for effectively discriminating between positive and negative classes.

Therefore, we selected ProteinBERT’s global embeddings as the default feature representation for our final DBP classification model, as it offered the
most consistent and generalizable performance across varying dataset sizes.
Their ability to capture both short- and long-range dependencies in protein sequences provides a substantial advantage, particularly for datasets with
rich diversity and sufficient volume. Despite the occasional strengths of one-hot encoding in low-data scenarios, its performance deteriorates on larger datasets, limiting its applicability for real-world deployment.

\subsection{Comparison with state-of-the-art models}

To rigorously evaluate the effectiveness of our proposed model, we performed comprehensive comparisons against several state-of-the-art (SOTA) methods. We report both the 5-fold cross-validation results (TTable~\ref{comparison_training}, Figure~\ref{train_reliability}) on the training set and the independent test set results (Table~\ref{comparison_testing}, Figure~\ref{test_reliability}) to ensure fair and robust assessment.

Before presenting comparative results, it is essential to clarify a critical methodological refinement made during our evaluation. Specifically, we addressed methodological concerns regarding the five-fold cross-validation strategy employed in the original BiCaps-DBP paper~\cite{mursalim2023bicaps}. Specifically, we corrected their evaluation protocol by conducting model assessment using a true five-fold cross-validation scheme, in line with standard machine learning practices. Under this revised and more rigorous validation framework, the reimplemented BiCaps-DBP(C) model exhibited a notable decline in performance across multiple metrics. This degradation can be primarily attributed to the original study’s failure to isolate the test set from the training process, effectively inflating performance estimates due to data leakage.

\begingroup
\setlength{\tabcolsep}{4pt} 
\renewcommand{\arraystretch}{1.35} 
\begin{table}[t]
\centering
\caption{Performance comparison of 5 folds cross-validation results on training datasets}
\label{comparison_training}

\resizebox{\linewidth}{!}{%
    \begin{tabular}{c|c|c|c|c|c}
    \hline
        \hline
    Method & ACC. (\%) & SEN. (\%) & SPE. (\%) & PRE. (\%) & MCC.(\%) \\
       \hline
        \hline

    \multicolumn{6}{c}{PDB14189} \\
    \hline
        \hline
    DBP-CNN\cite{barukab2022dbp} & 83.09 & 80.66 & 85.54 & 78.9 & 66.26 \\
    Target-DBPPred\cite{ali2022target} & 86.96  & 86.59 & 86.82  & 81.2 & 71 \\
    BiCaps-DBP(C)\cite{mursalim2023bicaps} & 76.44 & \textbf{98.34} & 52.82 & 73.43 & 51.44 \\
    ResCap-DBP & \textbf{95.01} & 97.43 & \textbf{90.42} & \textbf{91.11} & \textbf{88.18} \\
    \hline
        \hline
    \multicolumn{6}{c}{PDB1075} \\
    \hline
        \hline
    DNAbinder\cite{raghava2007identification} & 73.58 & 66.47 & \textbf{80.36} & 72.4 & 47.00 \\
    iDNA-Prot\cite{lin2011idna} & 75.40 & 83.81 & 64.73 & 73.9 & 50 \\
    BiCaps-DBP(C)\cite{mursalim2023bicaps} & 78.88 & \textbf{97.3} & 49.74 & 68.88 & 50.38 \\
    ResCap-DBP & \textbf{84.65} & 94.49 & 66.31 & \textbf{75.41} & \textbf{63.51} \\
    \hline
        \hline
    \multicolumn{6}{c}{PDB67151} \\
    \hline
        \hline
    CNN--BiLSTM\cite{hu2019improved} & 94.88 & 94.88 & 94.87 & 96.0 & 89.75 \\
    BiCaps-DBP(C)\cite{mursalim2023bicaps} & 98.69 & \textbf{99.58} & 97.83 & 97.87 & 97.39 \\
    ResCap-DBP & \textbf{98.85} & \textbf{99.58} & \textbf{98.24} & \textbf{98.26} & \textbf{97.76} \\
    \hline
        \hline
    \multicolumn{6}{c}{UNIPROT1424} \\
      \hline
        \hline
    StackDPP~\cite{ahmed2024stackdpp} & \textbf{92.00} & 92.00 & \textbf{92.00} & 88.0 & \textbf{84.00} \\
    ResCap-DBP & 91.1 & \textbf{94.1} & 88.1 & \textbf{88.7} & 82.31 \\
    \hline
        \hline
\end{tabular}
}
\end{table}
\endgroup

        \begin{figure}[!t]
        \centering
        \includegraphics[width=\linewidth]{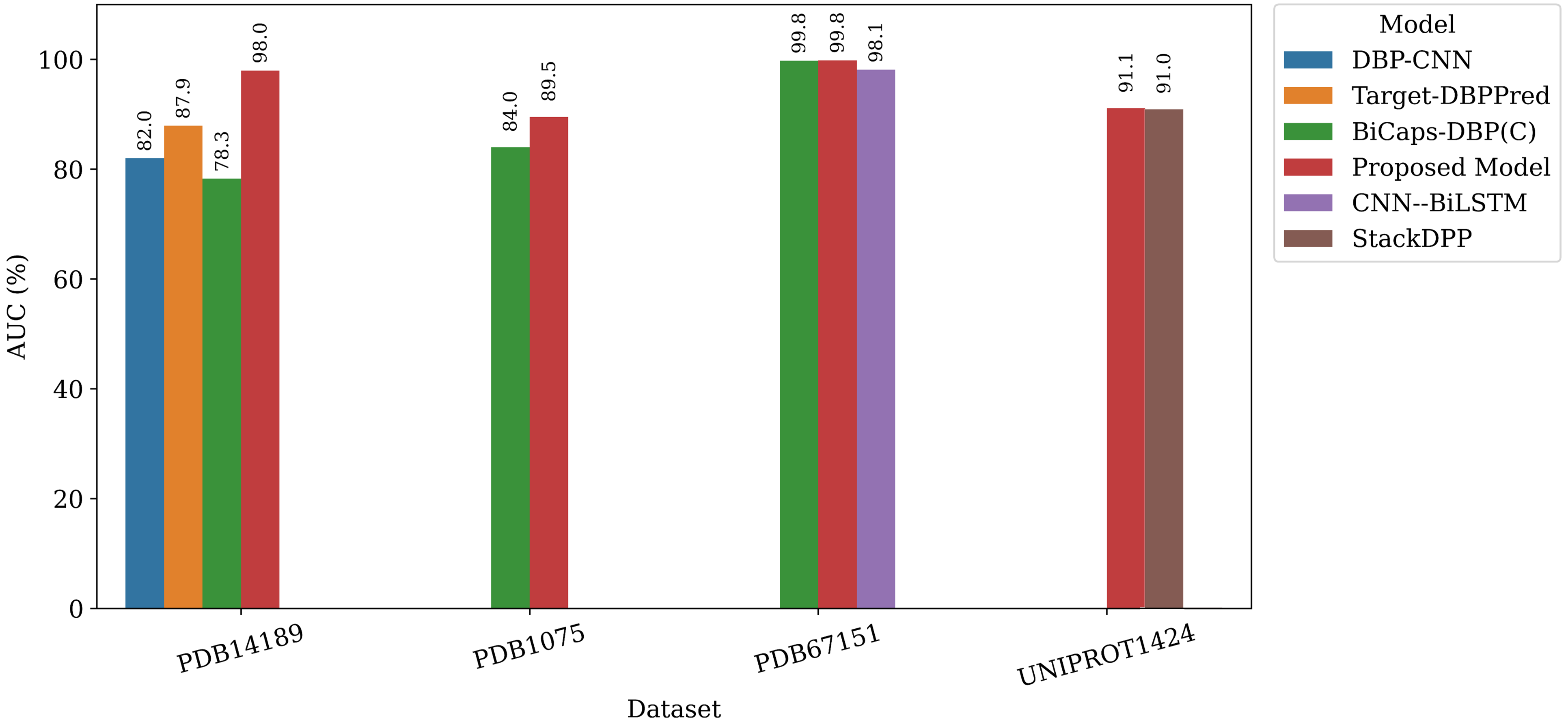}
        \caption{Comparison of the area under the curve (AUC) scores over 5-folds cross validation across different models on training datasets.}
        \label{train_reliability}
    \end{figure}
\par On the large-scale dataset PDB14189, the proposed model achieved the highest accuracy (95.01\%) and AUC (98.0\%), surpassing the best method (Target-DBPPred) by over 8\% in accuracy and over 10\% in AUC. Notably, it also recorded the highest specificity (90.42\%) and MCC (88.18\%), reflecting excellent balance and discriminative power. Although BiCaps- DBP(C) reported the highest sensitivity (98.34\%), its specificity dropped drastically (52.82\%), indicating a strong bias toward positive predictions and limited generalizability. On PDB1075, the proposed model continued to demonstrate leading performance, achieving the highest accuracy (84.65\%) and MCC (63.51\%), and outperforming BiCaps-DBP(C) and iDNA-Prot in both precision and specificity. While BiCaps-DBP(C) obtained the highest sensitivity (97.30\%), it again suffered from poor specificity (49.74\%), reinforcing the earlier observation of imbalanced predictions. On PDB67151, a relatively larger training dataset, our model achieved top scores across all metrics except sensitivity (which it shared with BiCaps-DBP(C) at 99.58\%). With an MCC of 97.76\% and an AUC of 99.8\%, it demonstrated near-perfect classification. These results highlight the model’s ability to scale with high-complexity data and retain accuracy and class balance. On UNIPROT1424, although StackDPP marginally outperformed the proposed model in accuracy (92.0\% vs. 91.1\%) and MCC (84.0\% vs. 82.31\%), our model showed superior sensitivity (94.1\% vs. 92.0\%) and precision (88.7\%), demonstrating a favorable balance and reliability, especially in positive-class recognition, which is critical in DBP identification.

\par Testing datasets pose a more stringent challenge, as they assess model generalization to unseen data. On PDB2272, although Target-DBPPred attained the highest accuracy (82.06\%) and MCC (63.00\%), the proposed model exhibited competitive performance (79.29\% accuracy, 50.21\% MCC), and reported the highest precision (70.76\%). In contrast, BiCaps-DBP(C) achieved high sensitivity (90.86\%) but performed poorly in specificity (39.03\%) and MCC (30.57\%), indicating overfitting to the positive class. For the small-scale PDB186 dataset, the proposed model achieved the highest accuracy (67.71\%), specificity (46.41\%), and precision (61.66\%), with a competitive MCC (34.40\%) that closely approached the best performer (DNA-BIND, 35.50\%). BiCaps-DBP(C), despite high sensitivity (89.68\%), again suffered from low specificity (36.11\%) and MCC (28.42\%). On the large-scale PDB20000 dataset, the proposed model achieved the best overall results, with the highest accuracy (89.78\%), specificity (86.41\%), precision (87.21\%), and MCC (79.22\%), while maintaining the same sensitivity (92.59\%) as BiCaps-DBP(C). The corresponding AUC of 96.2\% further confirms the model’s superior generalization on large, heterogeneous sequences. On UNIPROT356, the best MCC (86.00\%) and specificity (96.00\%) were achieved by StackDPP. However, our model demonstrated a significantly higher sensitivity (98.95\%) and nearly identical precision (87.04\%), confirming its robustness in detecting true positive instances. With an MCC of 85.77\% and an AUC of 92.8\%, the proposed model remains highly competitive.
Overall, the proposed model demonstrates state-of-the-art performance across a wide range of datasets, showing clear advantages in sensitivity, precision, and AUC. Its robust generalization and consistent performance across diverse dataset sizes and characteristics demonstrate its suitability for real-world DBP classification tasks.

\begingroup
\setlength{\tabcolsep}{4pt} 
\renewcommand{\arraystretch}{1.35} 
\begin{table}[!t]
\centering
    \caption{Performance comparison on testing datasets }
    \label{comparison_testing}

\resizebox{\linewidth}{!}{%
    \begin{tabular}{c|c|c|c|c|c}
    \hline
        \hline
    Method & ACC. (\%) & SEN. (\%) & SPE. (\%) & PRE. (\%) & MCC. (\%) \\
    \hline
        \hline
    \multicolumn{6}{c}{PDB2272} \\
    \hline
        \hline
    DBP-CNN\cite{barukab2022dbp} & 67.91 & 69.04 & 66.76 & 67.50 & 35.80 \\
    Target-DBPPred\cite{ali2022target} & \textbf{82.06} & 87.10 & \textbf{76.78} & 78.95 & \textbf{63.00} \\
    BiCaps-DBP(C)\cite{mursalim2023bicaps} & 66.13 & \textbf{90.86} & 39.03 & 62.49 & 30.57 \\
    ResCap-DBP & 79.29 & 83.99 & 65.14 & \textbf{70.76} & 50.21 \\
    \hline
        \hline
    \multicolumn{6}{c}{PDB186} \\
    \hline
        \hline
    DNA-BIND\cite{szilagyi2006efficient} & 67.70 & 66.70 & \textbf{68.80} & 68.13 & \textbf{35.50} \\
    iDNA-Prot\cite{lin2011idna} & 67.20 & 67.70 & 66.70 & 67.00 & 34.36 \\
    BiCaps-DBP(C)\cite{mursalim2023bicaps} & 66.67 & \textbf{89.68} & 36.11 & 59.43 & 28.42 \\
    ResCap-DBP & \textbf{67.71} & 84.94 & 46.41 & \textbf{61.66} & 34.4 \\
    \hline    \hline
        \hline    \multicolumn{6}{c}{PDB20000} \\
    \hline
        \hline
    CNN--BiLSTM\cite{hu2019improved} & 82.06 & 87.10 & 76.78 & 78.95 & 63.00 \\
    BiCaps-DBP(C)\cite{mursalim2023bicaps} & 89.30 & \textbf{92.59} & 85.41 & 86.41 & 78.23 \\
    ResCap-DBP & \textbf{89.78} & \textbf{92.59} & \textbf{86.41} & \textbf{87.21} & \textbf{79.22} \\
    \hline
        \hline
    \multicolumn{6}{c}{UNIPROT356} \\
    \hline
        \hline
    StackDPP~\cite{ahmed2024stackdpp} & 93.00 & 90.00 & \textbf{96.00} & \textbf{95.74} & \textbf{86.00} \\
    ResCap-DBP & 92.50 & \textbf{98.95} & 86.67 & 87.04 & 85.77 \\
    \hline
        \hline
\end{tabular}
}
\end{table}
\endgroup

    \subsection{Computational efficiency of the proposed network}
In this section, we highlight the computational efficiency of our proposed model, focusing on the number of trainable parameters and end-to-end inference time. In addition to achieving strong predictive performance, our network offers significant advantages in terms of resource utilization. Specifically, the architecture comprises 608,806 trainable parameters, enabling efficient memory usage and faster training cycles. Moreover, the average inference time per sequence is just 0.084777 seconds, making the model highly suitable for real-time applications and large-scale DNA-binding protein (DBP) prediction tasks, where both speed and efficiency are essential.

        \begin{figure}[!t]
        \centering
        \includegraphics[width=\linewidth]{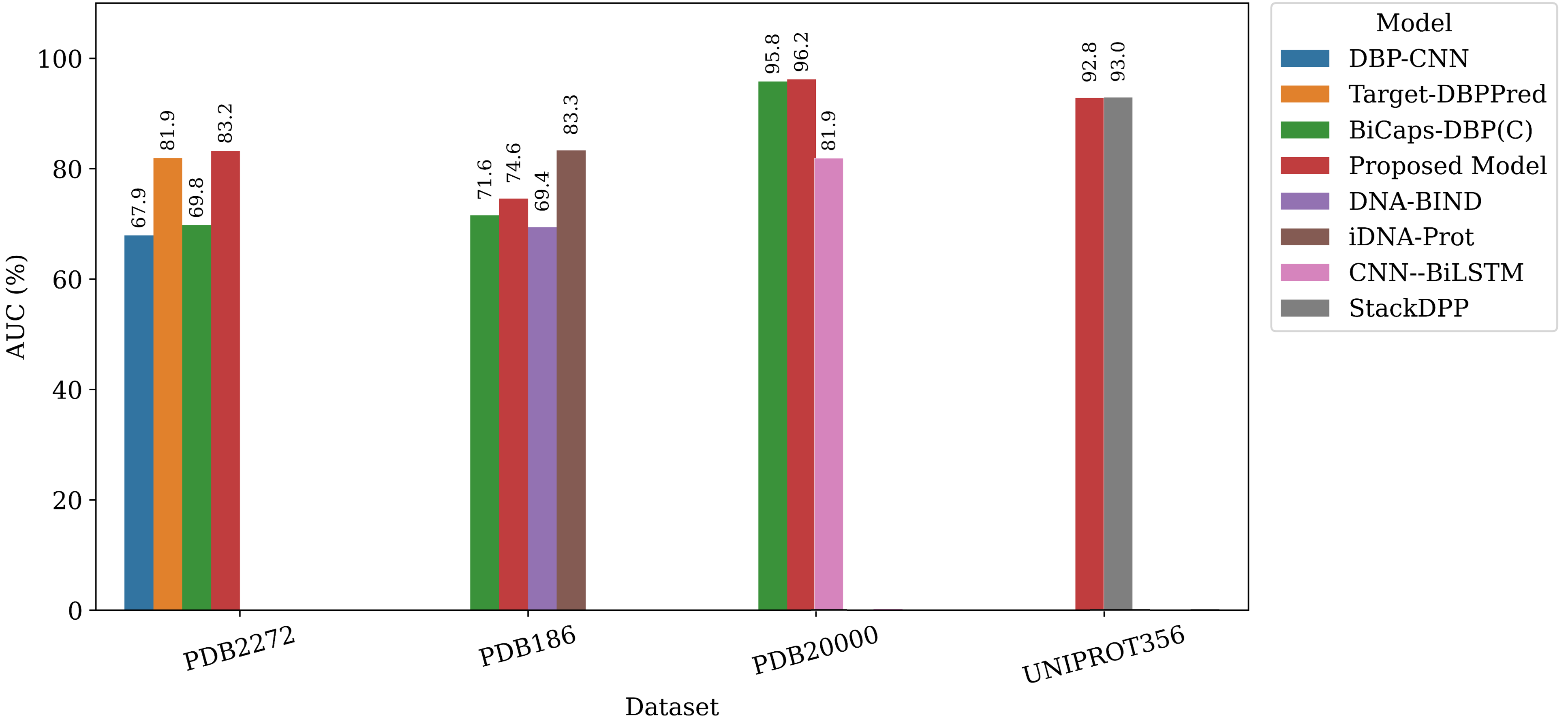}
        \caption{Comparison of the area under the curve (AUC) scores across different models on independent testing datasets.}
        \label{test_reliability}
    \end{figure}
\section{Future Works}\label{six}

Although ResCap-DBP has demonstrated strong predictive performance and computational efficiency, several directions remain to advance the classification of DNA-binding proteins further. Our critical analysis has exposed redundancy and high sequence similarity in widely used benchmark datasets. By coupling architectural innovation with a commitment to, we will undertake the construction of a new benchmark dataset that integrates structural and evolutionary evidence, such as three-dimensional protein coordinates and multiple-sequence alignments to enrich diversity and present more challenging evaluation scenarios and will provide a more solid foundation for DBP prediction, preventing potential false confidence situations and further advancing the field. Second, we plan to investigate the fusion of additional feature modalities, including predicted secondary structure profiles, physicochemical property indices, and genomic context signals, to augment our existing one-hot and ProteinBERT representations. These future efforts will culminate in the deployment of a user-friendly Web server and command line interface, enabling high-throughput genome-wide annotation of DNA-binding proteins in diverse organisms.

\section{Conclusion}\label{seven}

We have presented a lightweight, novel deep learning framework that synergistically combines residual dilated encoding with a 1D capsule network classifier for robust prediction of DNA-binding proteins. Through extensive experiments on multiple benchmark datasets, our model achieves state-of-the-art accuracy, sensitivity, and computational efficiency. By coupling architectural innovation with a commitment to data integrity, this work advances both the methodological toolkit and the practical evaluation standards for classifying DNA-binding proteins. We hope our work advances DBP prediction and provides valuable insights for future research and applications. 

\section{Compliance with Ethical Standards}

\textbf{Funding:} None.

\vspace{6pt}

\textbf{Use of Generative AI:} Artificial intelligence tools were used solely to improve the clarity, grammar, and linguistic expression of this manuscript. These tools did not contribute to the generation of scientific content, data interpretation, or analysis. The authors remain fully responsible for the originality, accuracy, and integrity of all intellectual content presented.

\vspace{4pt}

\textbf{Conflict of Interest:} The author confirms there are no financial or non-financial interests that could have influenced the research presented in this manuscript.

\vspace{4pt}

\textbf{Registration of Clinical Trial:} Not applicable.

\vspace{4pt}

\textbf{Consent to Publish:} Not applicable.

\vspace{4pt}

\textbf{Consent to Participate:} Not applicable.

\vspace{4pt}

\textbf{Ethical Approval:} This research did not involve any experiments on humans or animals conducted by the author.


\vspace{4pt}

\textbf{Informed Consent:} Not applicable.

 \bibliographystyle{elsarticle-num} 
 \bibliography{ref}

\end{document}